\documentclass{article}
\usepackage{ijcai22}

\usepackage{times}
\usepackage{soul}
\usepackage{url}
\usepackage[hidelinks]{hyperref}
\usepackage[utf8]{inputenc}
\usepackage[small]{caption}
\usepackage{graphicx}
\usepackage{amsmath}
\usepackage{amsthm}
\usepackage{booktabs}
\usepackage{algorithm}
\usepackage{algorithmic}
\usepackage{multirow}
\usepackage{footnote}
\usepackage{latexsym}

\title{D-DPCC: Deep Dynamic Point Cloud Compression via 3D Motion Prediction}

\author{
Tingyu Fan\footnote{Equal contribution.}$^1$
\and
 Linyao Gao$^{*1}$\and
Yiling Xu$^1$\and
Zhu Li$^2$\And
Dong Wang$^3$
\affiliations
$^1$Cooperative Medianet Innovation Center, Shanghai Jiao Tong University\\
$^2$University of Missouri, Kansas City\\
$^3$Guangdong OPPO Mobile Telecommunications Corp., Ltd.\\
\emails
\{woshiyizhishapaozi, linyaog, yl.xu\}@sjtu.edu.cn,
zhu.li@ieee.org,
wangdong7@oppo.com
}

\begin{document}
\maketitle

\begin{abstract}
The non-uniformly distributed nature of the 3D dynamic point cloud (DPC) brings significant challenges to its high-efficient inter-frame compression. This paper proposes a novel 3D sparse convolution-based Deep Dynamic Point Cloud Compression (D-DPCC) network to compensate and compress the DPC geometry with 3D motion estimation and motion compensation in the feature space. In the proposed D-DPCC network, we design a {\it Multi-scale Motion Fusion} (MMF) module to accurately estimate the 3D optical flow between the feature representations of adjacent point cloud frames. Specifically, we utilize a 3D sparse convolution-based encoder to obtain the latent representation for motion estimation in the feature space and introduce the proposed MMF module for fused 3D motion embedding. Besides, for motion compensation, we propose a 3D {\it Adaptively Weighted Interpolation} (3DAWI) algorithm with a penalty coefficient to adaptively decrease the impact of distant neighbors. We compress the motion embedding and the residual with a lossy autoencoder-based network. To our knowledge, this paper is the first work proposing an end-to-end deep dynamic point cloud compression framework. The experimental result shows that the proposed D-DPCC framework achieves an average 76\% BD-Rate (Bjontegaard Delta Rate) gains against state-of-the-art Video-based Point Cloud Compression (V-PCC) v13 in inter mode.

\end{abstract}

\section{Introduction}
In recent years, the dynamic point cloud (DPC) has become a promising data format for representing sequences of 3D objects and scenes, with broad applications in AR/VR, autonomous driving, and robotic sensing \cite{zhang2021attan}. However, compared with pixelized 2D image/video, the non-uniform distribution of DPC makes the exploration of temporal correlation extremely difficult, bringing significant challenges to its inter-frame compression. This paper focuses on the dynamic point cloud geometry compression with 3D motion estimation and motion compensation to reduce temporal redundancies of 3D point cloud sequences.

The existing dynamic point cloud compression (DPCC) methods can be concluded as 2D-video-based and 3D-model-based methods \cite{Li21.pcpc}. For 2D-video-based DPCC, the Moving Picture Expert Group (MPEG) proposes Video-based Point Cloud Compression (V-PCC) \cite{schwarz2018emerging}, which projects DPC into 2D geometry and texture video and uses mature video codecs (e.g., HEVC) for high-efficient video compression. Among all DPCC methods, V-PCC achieves current state-of-the-art performance. On the other hand, the 3D-model-based methods rely on motion estimation and motion compensation on 3D volumetric models like octree \cite{de2017motion}. However, the above methods are rule-based, with hand-crafted feature extraction modules and assumption-based matching rules, resulting in unsatisfactory coding efficiency.

Recently, the end-to-end learnt static point cloud (SPC) compression methods have reported remarkable gains against traditional SPC codecs like Geometry-based PCC (G-PCC) and V-PCC (intra) \cite{xu2018introduction}. Most of the learnt SPC compression methods are built upon the autoencoder architecture for dense object point clouds \cite{wang2021lossy,gao2021point,wang2021multiscale}, which divide SPC compression into three consecutive steps: feature extraction, deep entropy coding, and point cloud reconstruction. However, it is non-trivial to migrate the SPC compression networks to DPC directly. The critical challenge is to embed the motion estimation and motion compensation into the end-to-end compression network to remove temporal redundancies.

To this end, we propose a Deep Dynamic Point Cloud Compression (D-DPCC) framework, which optimizes the motion estimation, motion compensation, motion compression, and residual compression module in an end-to-end manner. Our contributions are summarized as follows:

\begin{enumerate}
    \item We first propose an end-to-end Deep Dynamic Point Cloud Compression framework (D-DPCC) for the joint optimization of motion estimation, motion compensation, motion compression, and residual compression.
    \item We propose a novel Multi-scale Motion Fusion (MMF) module for point cloud inter-frame prediction, which extracts and fuses the motion flow information at different scales for accurate motion estimation.
    \item For motion compensation, we propose a novel 3D Adaptively Weighted Interpolation (3DAWI) algorithm, which utilizes the neighbor information and adaptively decreases the impact of distant neighbors to produce a point-wise prediction of the current frame's feature. 
    \item Experimental result shows that our framework outperforms the state-of-the-art V-PCC v13 by an average 76\% Bjontegaard Delta Rate (BD-Rate)  gain on the 8iVFB \cite{d20178i} dataset suggested by MPEG.
\end{enumerate}

\section{Related Work}
\paragraph{Dynamic Point Cloud Compression.}
Existing DPCC works can be mainly summarized as 2D-video-based and 3D-model-based methods. The 2D-video-based methods perform 3D-to-2D projection to utilize the 2D motion estimation (ME) and motion compensation (MC) algorithms. Among them, MPEG V-PCC \cite{schwarz2018emerging} reports state-of-the-art performance. On the contrary, the 3D-model-based methods directly process the original 3D DPC. \cite{thanou2015graph} 
adopts geometry-based graph matching for attribute coding. \cite{de2017motion} breaks the DPC into blocks of voxels and performs block-based matching for 3D ME and MC to encode the point cloud geometry and attribute. However, the 3D-model-based methods fail to fully exploit the temporal correlations, leading to inferior results than 2D-video-based methods.

\paragraph{Learnt Static Point Cloud Compression.}
Inspired by the implementation of deep learning techniques in image/video compression, recently learnt static point cloud compression (SPCC) networks have emerged. \cite{huang20193d,gao2021point} design point-based SPCC networks to compress the 3D point clouds directly. However, these works are limited to small-scale point cloud datasets with only thousands of points. Wang et al. \cite{wang2021lossy} utilize 3D CNN-based frameworks for voxelized point cloud compression, achieving remarkable compression efficiency at the cost of excessive time and memory complexity. Thanks to the development of 3D sparse convolution \cite{choy20194d}, Wang et al. \cite{wang2021multiscale} propose an end-to-end sparse convolution-based multi-scale SPCC network, which reports state-of-the-art performance on large-scale datasets.

\paragraph{Point Cloud Scene Flow Estimation.}
Recently, 3D scene flow estimation has received much attention, which inspires us to compress DPC geometry by 3D motion estimation. FlowNet3D \cite{liu2019flownet3d} is the representative work of 3D scene flow estimation, which integrates a set conv layer based on PointNet++ \cite{qi2017pointnet++} for feature extraction, followed by a flow embedding layer for point matching. HALFlow \cite{wang2021hierarchical} proposes a double-attentive flow-embedding layer, which learns the correlation weights from both relative coordinates and feature space. Res3DSF \cite{wang2021residual} proposes a context-aware set conv layer and a residual flow learning structure for repetitive pattern recognition and long-distance motion estimation, which reports state-of-the-art performance. However, the above methods are trained on manually labelled scene flow data, leading to inefficient motion estimation for compression tasks. Furthermore, the above methods suffer from excessive time complexity, which are inapplicable on large-scale DPCs with hundreds of thousands of points in each frame.

\paragraph{Learnt Video Compression.}
There are massive attempts to apply deep learning techniques to video compression and build an end-to-end framework. Among them, DVC \cite{lu2019dvc} is the first video compression deep model that jointly optimizes all the components for video compression. For more accurate motion estimation, \cite{hu2021fvc} further proposes FVC by performing inter-frame prediction in the feature space, reporting superior performance over DVC.

\section{Methodology}
\begin{figure}[t]
\centering{
\includegraphics[width=0.925\linewidth]{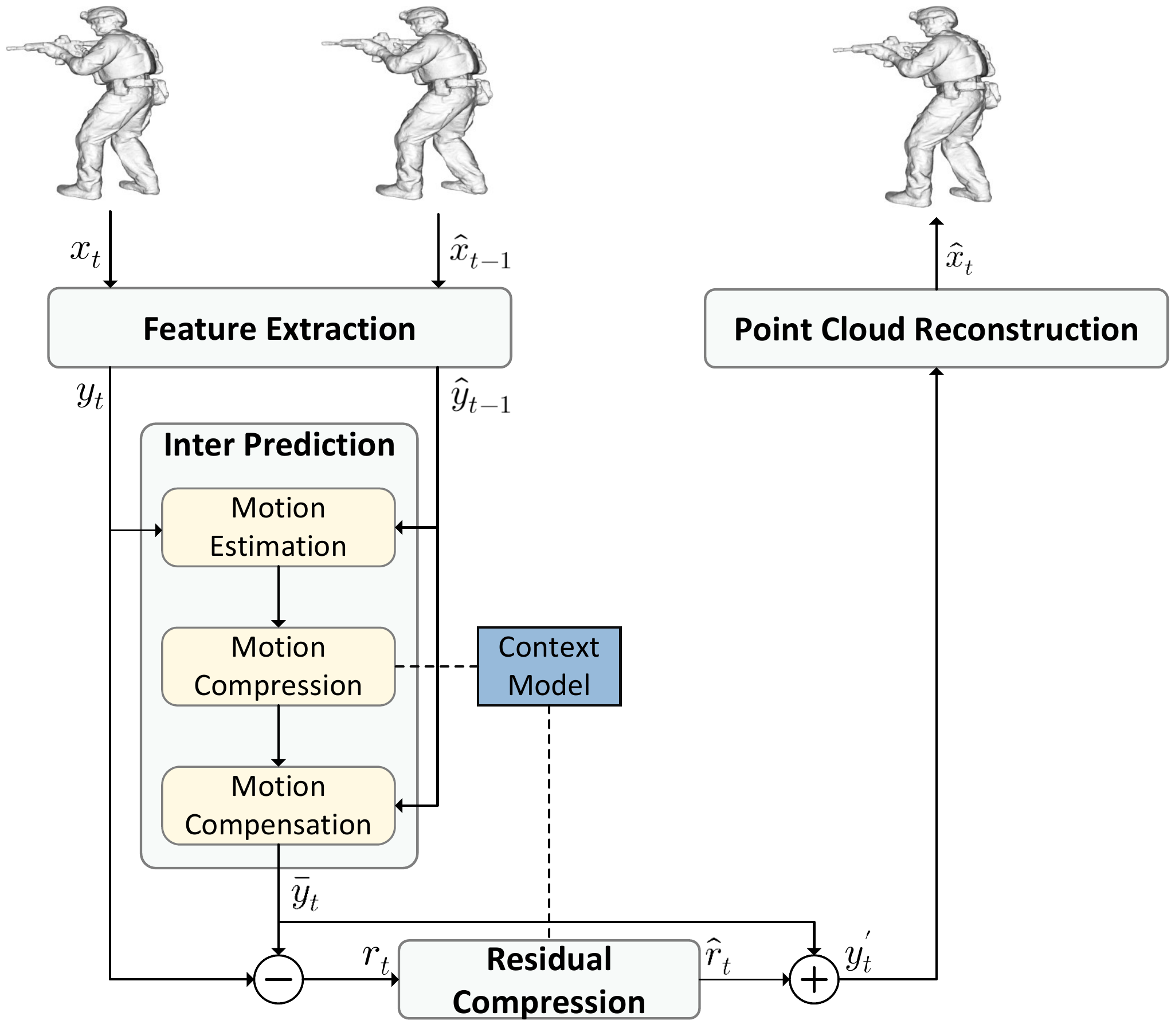}} 
\caption{The overall architecture of D-DPCC. $x_t$ and $\hat{x}_{t-1}$ are the current frame and the previously reconstructed frame. $y_t$ and $\hat{y}_{t-1}$ are the associated latent representations in feature space. $\bar{y}_t$ is the prediction of $y_t$. $r_t$ and $\hat{r}_t$ are the feature residual and reconstructed residual. $y_t'$ is the reconstruction value of $y_t$ in decoder. }
\label{overall}
\end{figure}

\begin{figure}[t]
\centering{
\includegraphics[width=0.925\linewidth]{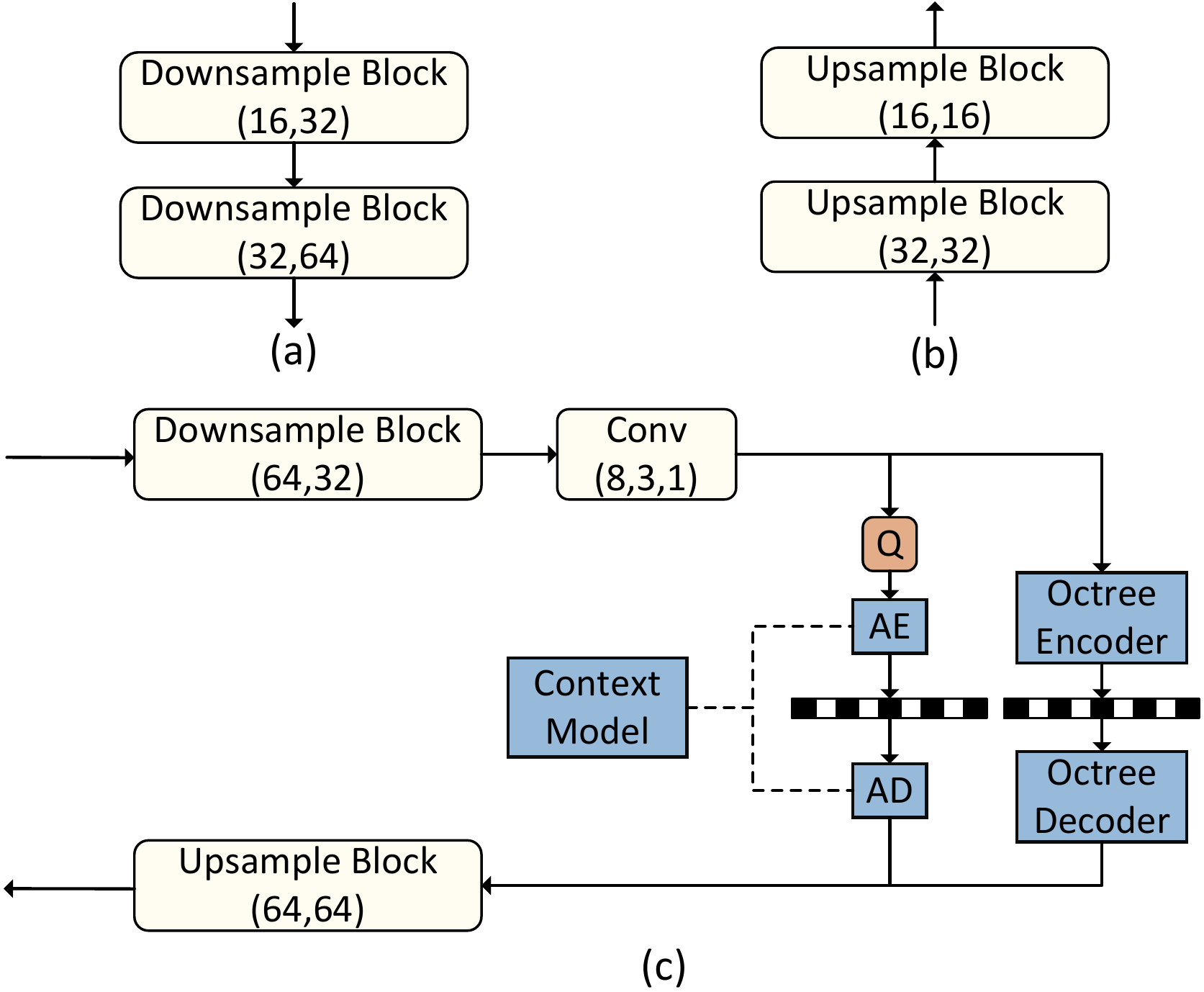}} 
\caption{The network architecture of (a) Feature Extraction, (b) Point Cloud Reconstruction, (c) Residual Compression.}
\label{overall_components}
\end{figure}

\subsection{Overview}
This section introduces the proposed end-to-end dynamic point cloud compression framework D-DPCC. The overall architecture of D-DPCC is shown in Figure \ref{overall}. Let $x_t=\{C_{x_t},F_{x_t}\}$ and $x_{t-1}=\{C_{x_{t-1}},F_{x_{t-1}}\}$ be two adjacent point cloud frames, where $C_{x_t}$ and $C_{x_{t-1}}$ are coordinate matrices. $F_{x_t}$ and $F_{x_{t-1}}$ are the associated feature matrices with all-one vectors to indicate voxel occupancy. The network analyses the correlation between $x_t$ and the previously reconstructed frame $\hat{x}_{t-1}$, aiming to reduce the bit rate consumption with inter-frame prediction. Specifically, $x_t$ and $\hat{x}_{t-1}$ are first encoded into latent representation $y_t$ and $\hat{y}_{t-1}$ in the feature extraction module. Then $y_t$ and $\hat{y}_{t-1}$ are fed into the inter prediction module to generate the predicted latent representation $\bar{y}_t$ of the current frame. The residual compression module compresses the feature residual $r_t$ between $y_t$ and $\bar{y}_t$. Finally, the reconstructed residual $\hat{r}_t$ is summed up with $\hat{y}_t$ and passes through the point cloud reconstruction module to produce the reconstruction $\hat{x}_t$ of the current frame.

\subsection{Feature Extraction}
\label{fe}
\begin{figure}[t]
\centering{
\includegraphics[width=0.925\linewidth]{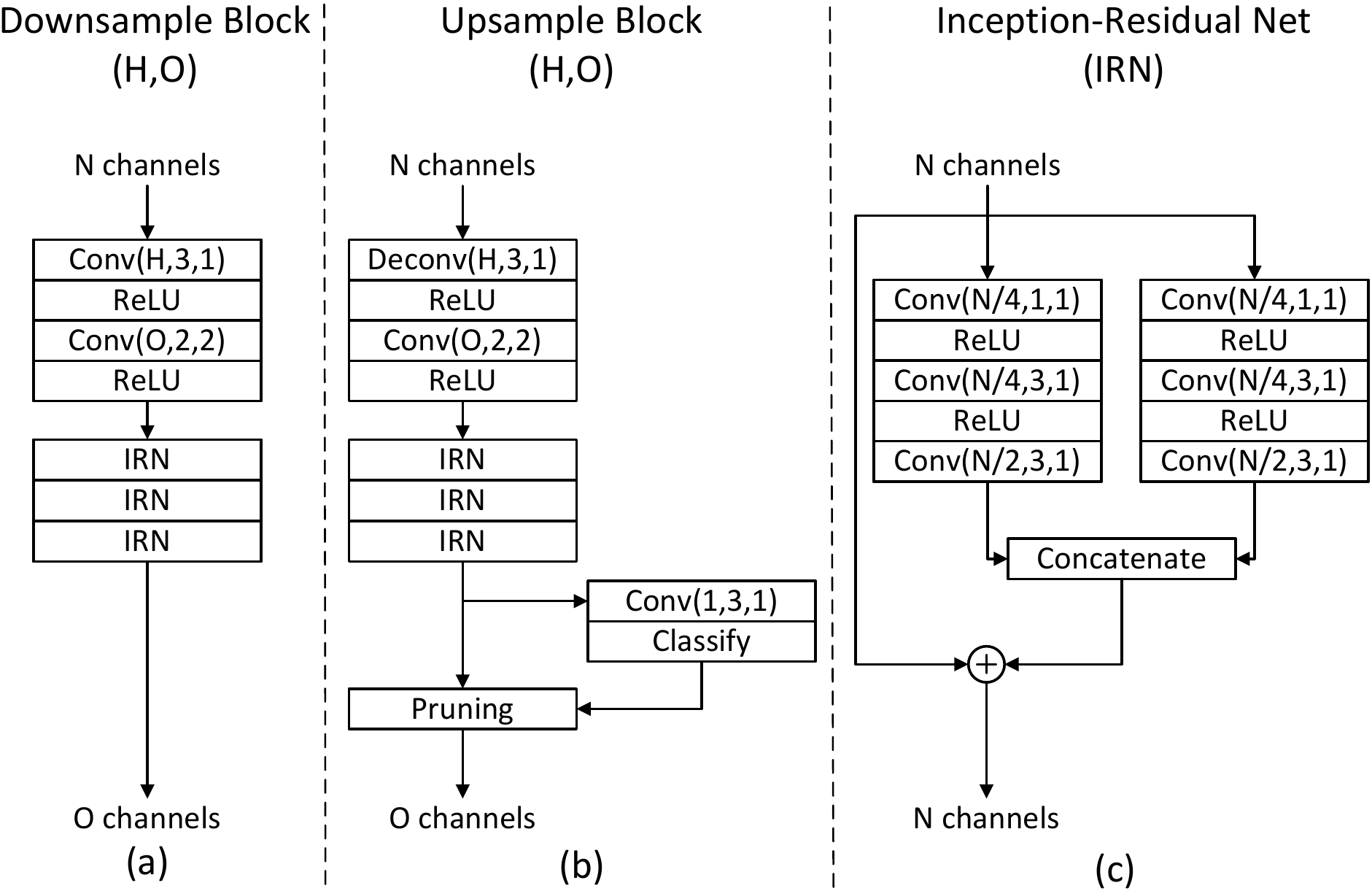}} 
\caption{The detailed architecture of each block. (a) Downsample Block, (b) Upsample Block, (c) Inception-Residual Block Net (IRN) block. $H$ and $O$ indicates the number of hidden and output channels.}
\label{components}
\end{figure}

The feature extraction module (Figure \ref{overall_components}(a)) consists of two serially connected Downsample Blocks for the hierarchical reduction of spatial redundancies, which encodes the current frame $x_t$ and the previously reconstructed frame $\hat{x}_{t-1}$ as latent representation $y_t$ and $\hat{y}_{t-1}$. Inspired by \cite{wang2021multiscale}, we adopt the sparse CNN-based Downsample Block (Figure \ref{components}(a)) for low-complexity point cloud downsampling. The Downsample Block consists of a stride-two sparse convolution layer for point cloud downsampling, followed by several Inception-Residual Network (IRN) blocks \cite{szegedy2017inception} (Figure \ref{components}(c)) for local feature analysis and aggregation. For simplicity of notations, we denote the coordinate matrix of $x_t$ at different downsampling scales as $C_{x_t}^k$, where $k$ is the scale index (i.e., $\times\frac{1}{2^k}$). Correspondingly, $y_t=\{C_{x_t}^2,F_{y_t}\}$, where $F_{y_t}$ is the feature matrix of $y_t$.

\subsection{Inter Prediction}
The inter prediction module takes the latent representation of both the current frame and the previously reconstructed frame, i.e., $y_t$ and $\hat{y}_{t-1}$ as input, analysing the temporal correlation and producing the feature prediction of $y_t$, i.e., $\bar{y}_t$.

The existing 3D scene flow estimation networks estimate a motion flow between two frames $p_1$ and $p_2$: $\mathcal{D}=\{x_i,y_i,z_i,{\bf d}_i|i=1,\cdots,N_1\}$ to minimize the distortion between $\mathcal{D}$ and the ground truth $\mathcal{D}^*$ \cite{liu2019flownet3d}. However, without end-to-end optimization, the predicted motion flow is unsatisfactory for motion compensation and motion compression. Therefore, we borrow the idea of feature-space motion compensation \cite{hu2021fvc} from video compression to design the end-to-end inter prediction module.

\begin{figure}[t]
\centering{
\includegraphics[width=0.975\linewidth]{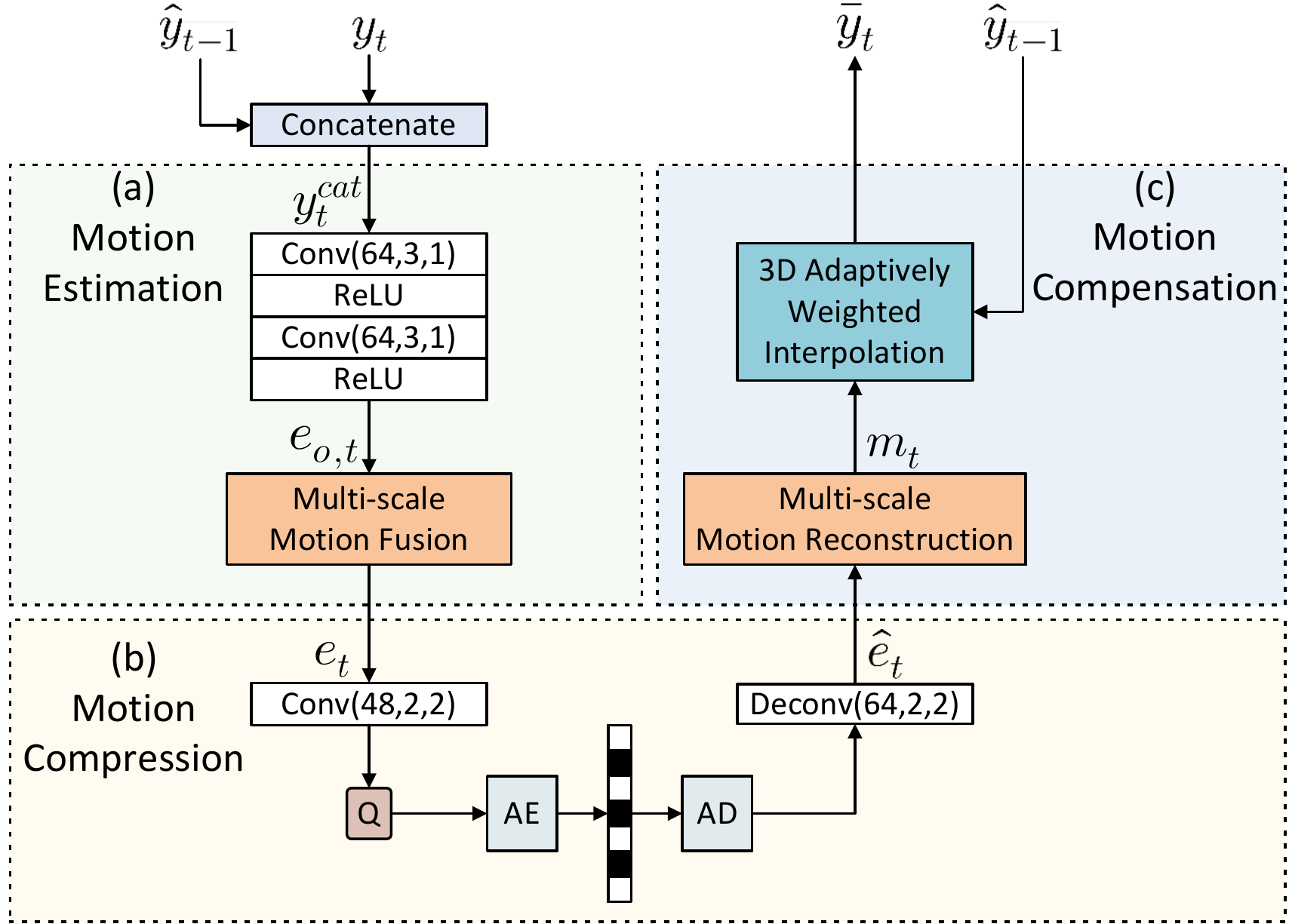}} 
\caption{The network architecture of the inter prediction module. (a) Motion estimation, (b) Motion compression, (c) Motion compensation. $e_{o,t}$ and $e_t$ are the original flow embedding and fused flow embedding. $\hat{e}_t$ is the reconstructed flow embedding. $m_t$ is the estimated motion flow.}
\label{interprediction}
\end{figure}

 The overall architecture of the inter prediction module is shown in Figure \ref{interprediction}. Specifically, we first {\it concatenate} $y_t$ and $\hat{y}_{t-1}$ together to get $y_t^{cat}$. The {\it concatenate} operation for point clouds is defined as:
\begin{equation}
x_u^{cat}=\left\{
\begin{array}{rcl}
x_{1,u}{\bf \oplus} x_{2, u} &            &  u\in C_1\cap C_2\\
{\bf 0}\oplus x_{2,u}       &      &  u\notin C_1\land u\in C_2\\
x_{1,u}\oplus {\bf 0}       &      & u\in C_1\land u\notin C_2\\

\end{array} \right. 
\end{equation}
where $x_u^{cat}$ is the concatenation of $x_1$ and $x_2$. $x_{1,u}$ and $x_{2,u}$ are the input feature vectors defined at $u$, and $\oplus$ is the concatenate operation between vectors. $y_t^{cat}$ passes through a 2-layer convolution network to generate the original flow embedding $e_{o,t}$. We further design a Multi-scale Motion Fusion (MMF) module for precise motion flow estimation. MMF extracts multi-scale flow embedding from $e_{o,t}$ and outputs a fused flow embedding $e_t$, which is encoded by an autoencoder-style network (Figure \ref{interprediction}(b)), whose encoder and decoder are each composed of a stride-two sparse convolution and transpose convolution layer, respectively.

\begin{figure}[t]
\centering{
\includegraphics[width=0.975\linewidth]{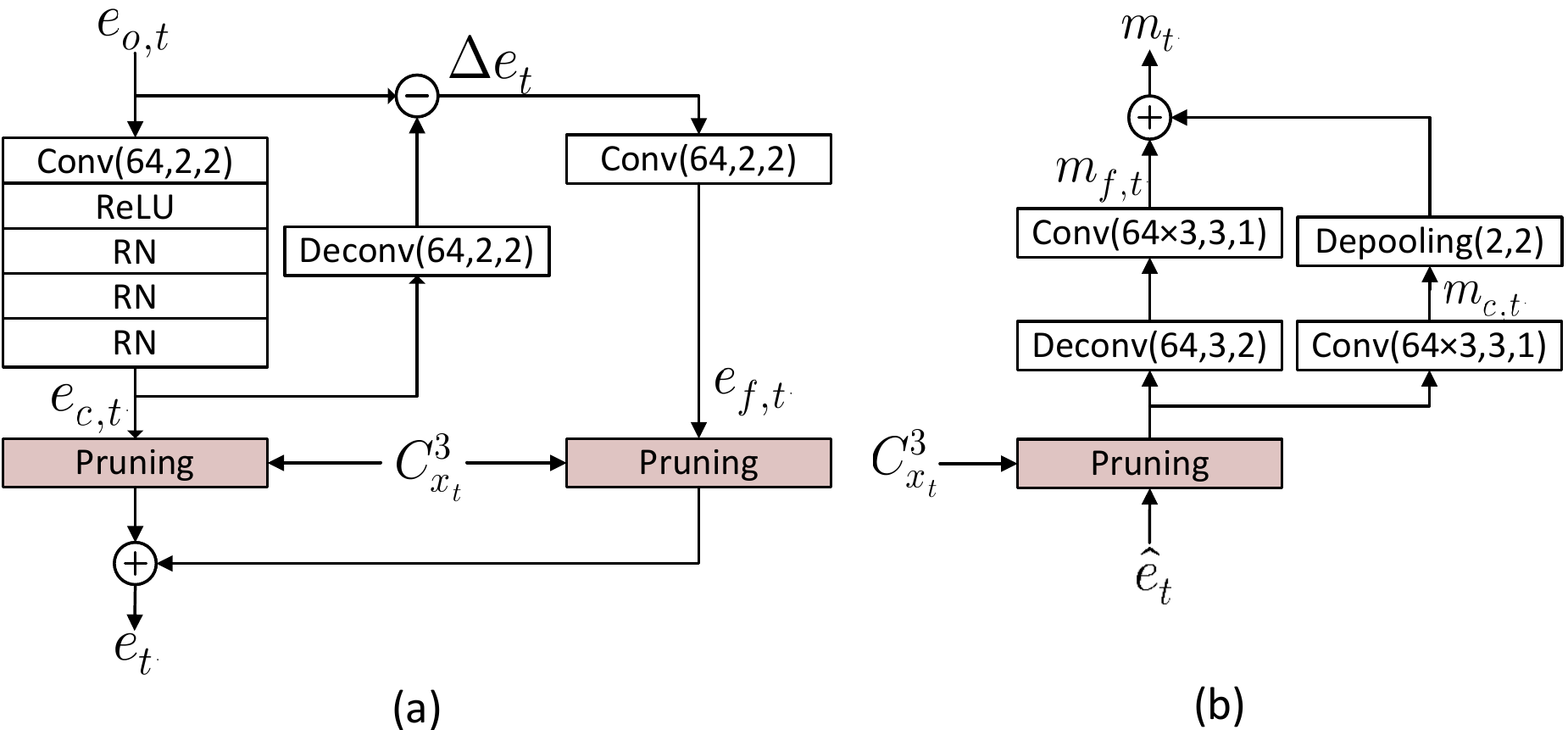}} 
\caption{The architecture of (a) Multi-scale Motion Fusion (MMF) module, (b) Multi-scale Motion Reconstruction (MMR) module.}
\label{MMF}
\end{figure}

\paragraph{Multi-scale Motion Fusion.}
Due to the sparsity nature of point clouds, the Euclidean distances between two adjacent frames' corresponding points have an enormous variance, leading to ineffective matching between points in $y_t$ and $\hat{y}_{t-1}$.  A straightforward solution is to increase the downsampling factor to expand the perception field. However, this leads to more information loss. Therefore, We propose a Multi-scale Motion Fusion (MMF) module as shown in Figure \ref{MMF}. The MMF module enhances the original flow embedding $e_{o,t}$ with the multi-scale flow information, which first downsamples $e_{o,t}$ with a stride-two sparse convolution layer to expand the perception field, then uses several Residual Net (RN) blocks to generate the coarse-grained flow embedding $e_{c,t}$. To compensate for the information loss during downsampling, we compute the residual (denoted as $\Delta e_t$) between upsampled $e_{c,t}$ and $e_{o,t}$, then downsample $\Delta e_t$ to obtain the fine-grained flow embedding $e_{f,t}$. Finally, $e_{c,t}$ is summed up with $e_{f,t}$ to generate the fused flow embedding $e_t$. 
Additionally, for multi-scale motion compensation, we design a Multi-scale Motion Reconstruction (MMR) module, which is symmetric to MMF. MMR recovers coarse-grained and fine-grained motion flow $m_{c,t}$ and $m_{f,t}$ from the reconstructed flow embedding $\hat{e}_t$, where $m_{c,t}$ is up-scaled and summed up with $m_{f,t}$ to generate the fused motion flow $m_t$.

\paragraph{3D Adaptively Weighted Interpolation.} 
The neural network estimates a continuous motion flow $m_t$, without the explicit specification to the corresponding geometry coordinates in the previous frame. Therefore, we propose a 3D Adaptively Weighted Interpolation (3DAWI) algorithm to obtain the derivable predicted latent representation $\bar{y}_t$ of the current frame. Given $m_t$ and $\hat{y}_{t-1}$, the predicted latent representation $\bar{y}_t$ is derived as follows:
\begin{equation}
\bar{y}_{t,u}=\frac{\sum_{v\in\vartheta_{u'}}d_{u',v}^{-1}\cdot\hat{y}_{t-1,v}}{{\rm max}(\sum_{v\in\vartheta_{u'}}d_{u',v}^{-1},\alpha)} {\ \rm for\ {\it u}\ in\ C_{x_t}^2},
\label{B3NI}
\end{equation}
where $\bar{y}_{t,u}$ is the predicted feature vector of $\bar{y}_t$ defined at $u$, $\hat{y}_{t-1,v}$ is the feature vector of $\hat{y}_{t-1}$ defined at $v$, $u'$ is $u$'s translated coordinate, $\vartheta_{u'}$ is the 3-nearest neighbors of $u'$ in $\hat{y}_{t-1}$, and $d_{u',v}$ is the squared Euclidean distance between coordinates $u'$ and $v$. $\alpha$ is a penalty coefficient related to point-spacing. Here, we intuitively give its value as $\alpha=3m$, where $m$ is the minimum point spacing in the dataset. The 3DAWI uses inverse distance weighted average (IDWA) based on geometric 3-nearest neighbors \cite{qi2017pointnet++}, which has a broader perception field than the trilinear interpolation algorithm. However, the IDWA based interpolation algorithm assigns an equal weighted sum of $1$ for each $u$ in $C_{x_t}^2$, inappropriate for $u$ with less effective corresponding points in $\hat{y}_{t-1}$. Therefore, 3DAWI utilizes $\alpha$ to adaptively decrease the weighted sum of each $u$ in $C_{x_t}^2$ whose translated coordinate $u'$ is {\it isolated} in $\hat{y}_{t-1}$, where $u'$ is {\it isolated} indicates that $\sum_{i\in\vartheta_{u'}}d_{u',i}^{-1}< \alpha$, i.e., $u'$ deviates from its 3-nearest neighbors, and the matching is less effective. Under extreme conditions, 
given $d_{u',v}\rightarrow\infty$ for each $v$ in $\vartheta_{u'}$, then $\bar{y}_{t,u}\rightarrow{\bf 0}$. Note that the coordinate matrix of $y_t$, i.e., $C_{x_t}^2$, is required in Equation (\ref{B3NI}), which is losslessly encoded \cite{nguyen2021learning} and transmitted to the decoder, and only accounts for a small amount ($<0.05$ bpp) of the whole bitstream. (See supplementary material for more details). 


\subsection{Residual Compression} 
The residual compression module encodes the residual $r_t$ between $y_t$ and $\bar{y}_t$ in feature space. We employ an autoencoder-style lossy compression network for residual compression as shown in Figure \ref{overall_components}(c). The encoder consists of a Downsample Block and a sparse convolutional layer, aiming to transform $r_t$ into latent residual $l_t=\{C_{x_t}^3,F_{l_t}\}$. $C_{x_t}^3$ is losslessly coded using G-PCC (octree) \cite{schwarz2018emerging}, and the feature matrix $F_{l_t}$ is quantized and coded with a factorized entropy model \cite{balle2017end}. The decoder uses an Upsample Block (Figure \ref{components}(b)) to recover the reconstructed residual $\hat{r}_t$, which is added with the predicted latent representation $\bar{y}_t$ to obtain the reconstructed feature $y'_t$ as shown in Figure \ref{overall}.

\subsection{Point Cloud Reconstruction}
The point cloud reconstruction module (Figure \ref{overall_components}(b)) mirrors the operations of feature extraction, with two serially connected Upsample Block (Figure \ref{components}(b)) for hierarchical reconstruction. The Upsample Block uses a stride-two sparse transpose convolution layer for point cloud upsampling. After successive convolutions, the Upsample Block uses a sparse convolution layer to generate the occupation probability of each voxel. To maintain the sparsity of the point cloud after upsampling, we employ adaptive pruning \cite{wang2021multiscale} to detach false voxels based on the occupation probability.

\subsection{Loss Function}
\label{lossfunc}
We apply the rate-distortion joint loss function for end-to-end optimization:
\begin{equation}
\mathcal{L}=\mathcal{R}+\lambda\mathcal{D},
\end{equation}
where $\mathcal{R}$ is the bits per point (bpp) for encoding the current frame $x_t$, and $\mathcal{D}$ is the distortion between $x_t$ and the decoded current frame $\hat{x}_t$.

\paragraph{Rate.}
The latent feature $f$ is quantized before encoding. Note that the quantization operation is non-differentiable, thus we approximate the quantization process by adding a uniform noise $\mu\sim\mathcal{U}(-0.5,0.5)$. The quantized latent feature $\bar{f}$ is encoded using arithmetic coding with a factorized entropy model \cite{balle2017end}, which estimates the probability distribution of $\bar{f}$, i.e., $p_{\tilde{f}|\psi}(\tilde{f}|\psi)$, where $\psi$ are the learnable parameters. Then the bpp of encoding $\tilde{f}$ is:
\begin{equation}
\mathcal{R}=\frac{1}{N}\sum_i\log_2p_{\tilde{f_i}|\psi^{(i)}}(\tilde{f_i}|\psi^{(i)}),
\end{equation}
where $N$ is the number of points, and $i$ is the index of channels.

\paragraph{Distortion.}
As shown in Figure \ref{components}(b), in the proposed Downsample Block, a sparse convolution layer is used to produce the occupation probability $p_v$ of each voxel $v$ in the decoded point cloud. Therefore, we apply binary cross entropy (BCE) loss to measure the distortion:
\begin{equation}
\mathcal{D}_{BCE}=\frac{1}{N}\sum_{v}-(\mathcal{O}_v\log p_v+(1-\mathcal{O}_v)\log (1-p_v))\text{,}
\end{equation}
where $\mathcal{O}_v$ is the ground truth that either $v$ is occupied (1) or unoccupied (0). For hierarchical reconstruction, the distortion of each scale is averaged, i.e.,

\begin{equation}
    \mathcal{D}=\frac{1}{K}\sum_{k=1}^K\mathcal{D}_{BCE}^{(k)},
\end{equation}
where $k$ is the scale index. 
\vspace{2mm}

\section{Experiments}
\subsection{Experimental Settings}
\paragraph{Training Dataset.} We train the proposed model using Owlii Dynamic Human DPC dataset \cite{keming2018owlii}, containing 4 sequences with 2400 frames. The frame rate is 30 frames per second (fps) over 20 seconds for each sequence. To reduce the time and memory consumption during training and exemplify the scalability of our model, we quantize the 11-bit precision point cloud data into 9-bit precision.

\paragraph{Evaluating Dataset.}
Following the MPEG common test condition (CTC), we evaluate the performance of the proposed D-DPCC framework using 8i Voxelized Full Bodies (8iVFB) \cite{d20178i}, containing 4 sequences with 1200 frames. The frame rate is 30 fps over 10 seconds.

\paragraph{Training Strategy.}
We train D-DPCC with $\lambda=$3, 4, 5, 7, 10 for each rate point. We utilize an Adam \cite{kingma2015adam} optimizer with $\beta=(0.9,0.999)$, together with a learning rate scheduler with a decay rate of 0.7 for every 15 epochs. A two-stage training strategy is applied for each rate point. Specifically, for the first five epochs, $\lambda$ is set as 20 to accelerate the convergence of the point cloud reconstruction module; then, the model is trained for another 45 epochs with $\lambda$ set to its original value. The batch size is 4 during training. We conduct all the experiments on a GeForce RTX 3090 GPU with
24GB memory.

\paragraph{Evaluation Metric.}
The bit rate is evaluated using bits per point (bpp), and the distortion is evaluated using point-to-point geometry (D1) Peak Signal-to-Noise Ratio (PSNR), and point-to-plane geometry (D2) PSNR following the MPEG CTC. The peak value $p$ is set as 1023 for 8iVFB.





\subsection{Experimental Results}
\begin{figure}[t]
\centering  
\includegraphics[width=0.5\textwidth]{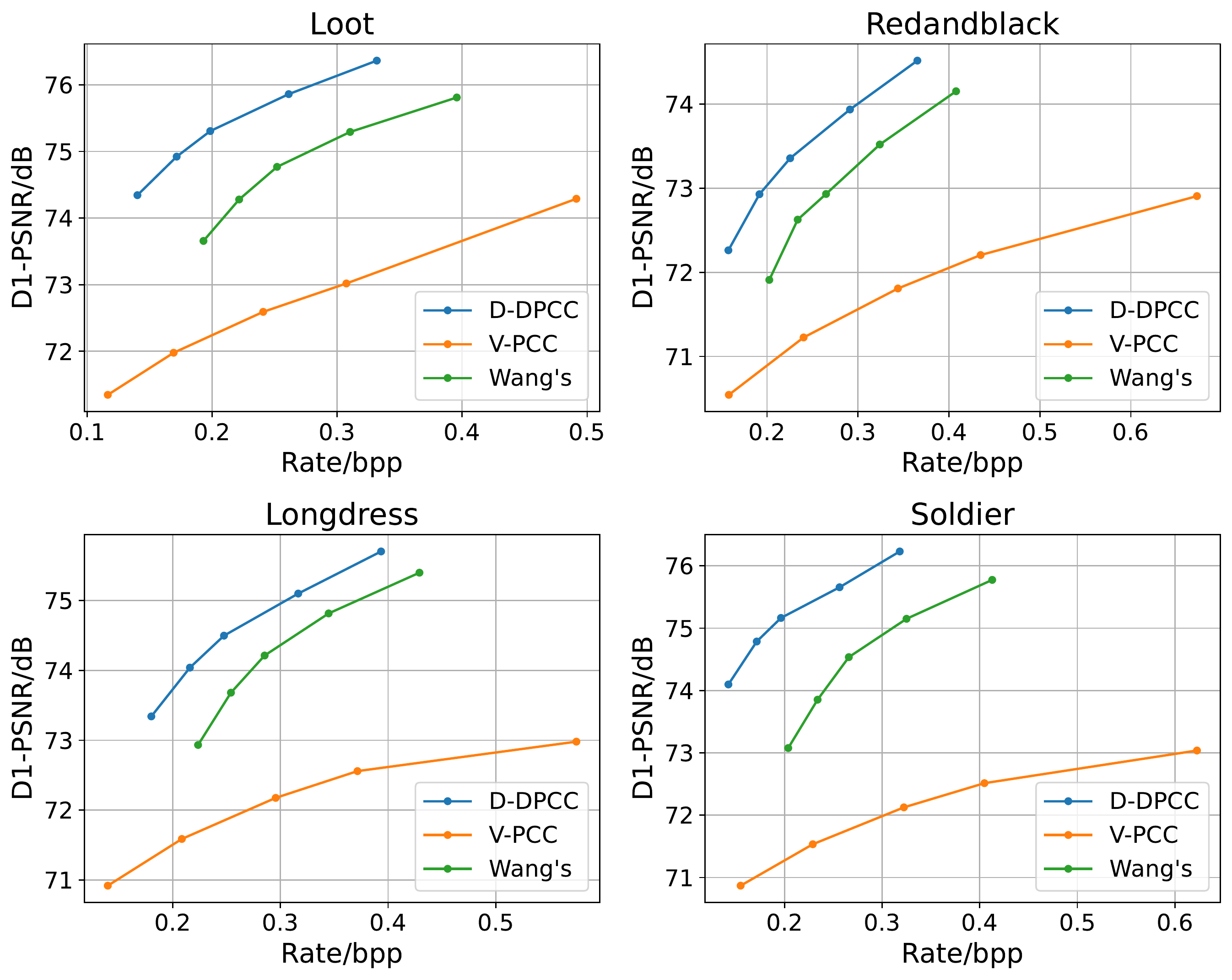}
\caption{D1 Rate-Distortion curves on 8iVFB (Loot, Redandblack, Longdress, Soldier) test sequences.}
\label{rd}
\end{figure}
  
\paragraph{Baseline Setup.} We compare the proposed D-DPCC with the current state-of-the-art dynamic point cloud geometry compression framework: V-PCC Test Model v13, with the quantization parameter (QP) setting as $18, 15, 12, 10, 8$, respectively. We also compare with Wang's framework \cite{wang2021multiscale}, which is state-of-the-art on static object point cloud geometry compression. For the fairness of comparison, we retrain Wang's framework using our training data and strategy, and the network parameters of each module are set the same as Wang's except for the proposed inter prediction module. When using the proposed D-DPCC for inter-frame coding, the first frame of the sequence is encoded using Wang's network with the same $\lambda$.
\begin{figure}[t]
\centering  
\includegraphics[width=0.4\textwidth]{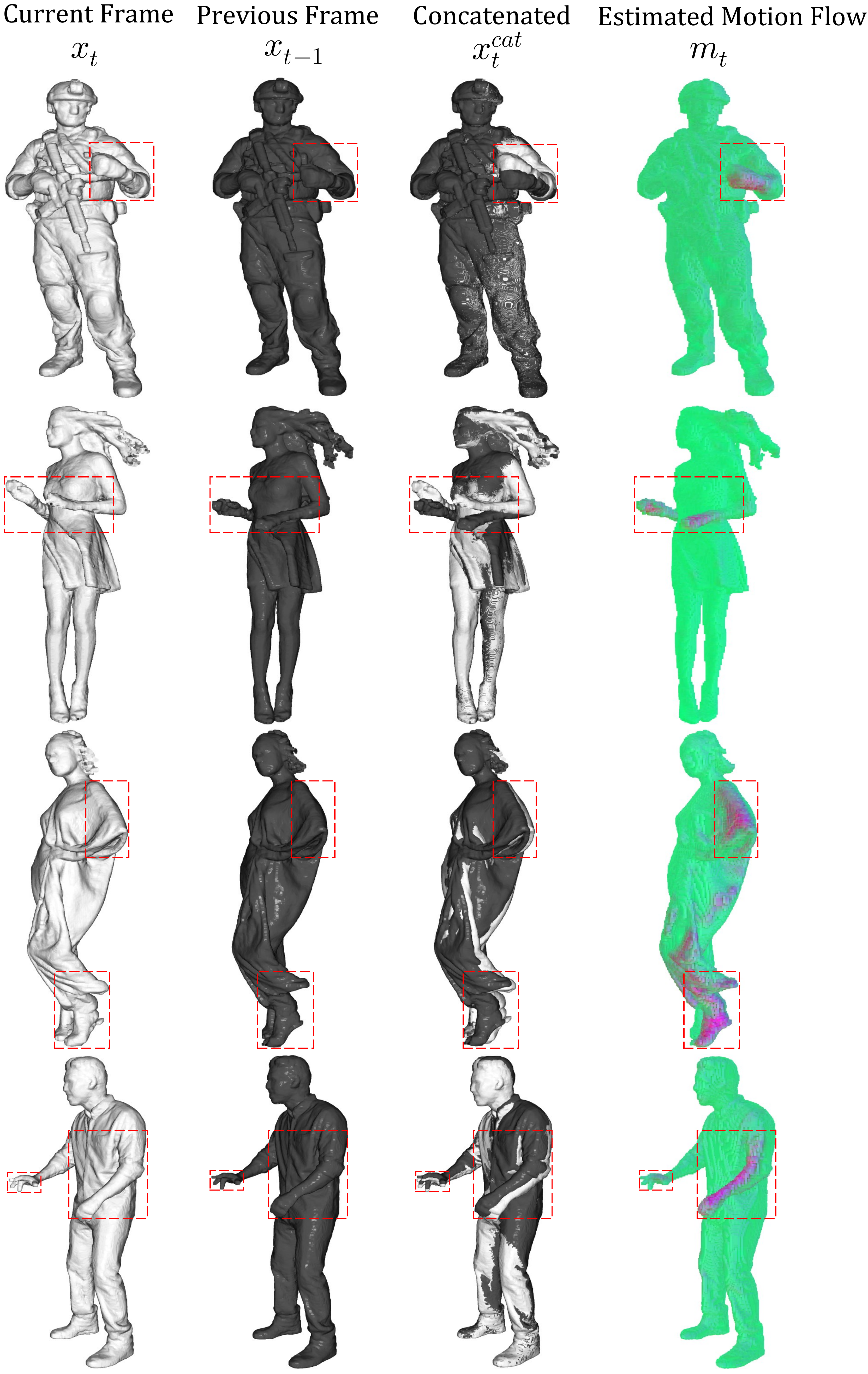}
\caption{Visualization of the estimated motion flow. The leftmost column is the current frames; the left middle column is the previous frames; the middle right column is the concatenations of the current frames and the previous frames. The rightmost column visualizes the predicted motion flow, with green indicating stillness and purple indicating movement. }
\label{flow}
\end{figure}

\begin{table}[t]
    \centering
    \begin{tabular}{cc cc cc}
      \toprule
      \multicolumn{2}{c}{Sequence} & \multicolumn{2}{c}{V-PCC} & \multicolumn{2}{c}{Wang's}  \\ 
      \multicolumn{2}{c}{ } & D1 & D2 & D1 & D2\\ 
      \midrule
      \multicolumn{2}{c}{Loot} & -71.76  & -70.92  & -36.15  & -29.73\\
      \multicolumn{2}{c}{Redandblack} & -68.97  & -76.71  & -25.68  & -20.56\\
      \multicolumn{2}{c}{Soldier} & -85.71  & -73.08  & -39.48  & -36.47\\
      \multicolumn{2}{c}{Longdress} & -78.92  & -72.89  & -22.31  & -17.91\\
      \midrule
      \multicolumn{2}{c}{\textbf{Overall}} & \textbf{-76.66}  & \textbf{-74.43}  & \textbf{-31.14}  & \textbf{-26.39}\\
      \bottomrule
     \end{tabular}
     \caption{BD-Rate(\%) gains against V-PCC (inter) and Wang's framework. For the proposed D-DPCC, the first frame of a sequence is encoded by Wang's network. The encoding of each subsequent frame takes the previous reconstructed frame as a reference.}
      \label{tab:bd}
\end{table}  

\paragraph{Performance Analysis.}
The rate-distortion curves of different methods are presented in Figure \ref{rd}, and the corresponding BD-Rate gains are shown in Table \ref{tab:bd}. Our D-DPCC outperforms V-PCC v13 by a large margin on all test sequences, with an average 76.66\% (D1), 74.43\% (D2) BD-Rate gain. Compared with Wang's network, the experimental result demonstrates that D-DPCC significantly improves the coding efficiency on test sequences Loot and Soldier with small motion amplitude, achieving $>36\%$ (D1), $>29\%$ (D2) BD-Rate gains. Meanwhile, D-DPCC still reports a considerable bit-rate reduction on Longdress and Redandblack with larger motion amplitude with $>22\%$ (D1), $>17\%$ (D2) BD-Rate gain. 
On all test sequences, D-DPCC achieves an average $31.14\%$ (D1) and $26.39\%$ (D2) BD-Rate gain against Wang's network. D-DPCC integrates a learnt inter prediction module with an MMF module for accurate motion estimation and the 3DAWI algorithm for point-wise motion compensation, which is jointly optimized with the other modules in D-DPCC. Therefore, D-DPCC significantly surpasses V-PCC (inter) and Wang's method.

\subsection{Analysis and Ablation Study}
\paragraph{Effectiveness of Inter Prediction.}
To verify the effectiveness of the inter prediction module, we visualize the estimated motion flow $m_t$ between two adjacent frames in Figure \ref{flow}. It can be observed that although no ground truth of 3D motion flow is provided during training, the inter prediction module still learns an explainable motion flow based on the analysis of movements between two adjacent point cloud frames through end-to-end training.

\begin{figure}[t]
\centering  
\includegraphics[width=0.455\textwidth]{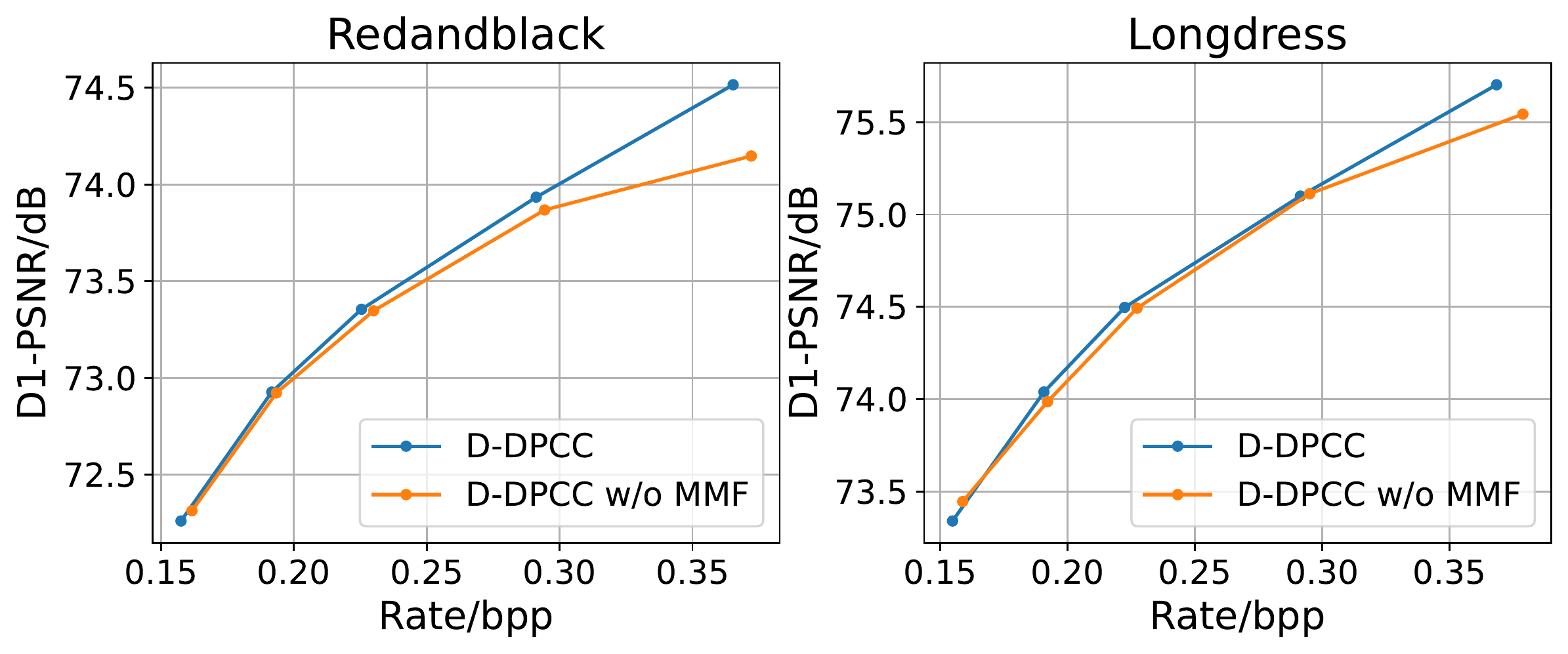}
\caption{Ablation study on Redandblack and Longdress for Multi-scale Motion Fusion (MMF) module.}
\label{MMF_result}
\end{figure}

\paragraph{Effectiveness of Multi-scale Motion-fusion.}
Figure \ref{MMF_result} reports the D1 rate-distortion curves of D-DPCC with/without the Multi-scale Motion Fusion (MMF) module. It is noted that with the multi-scale flow embedding and the fusion of the coarse-grained and fine-grained motion flow, MMF improves the coding efficiency by $>4\%$ BD-Rate gains. The improvement is particularly significant at high bit rates, where the network is encouraged to produce precise motion flow estimation. The performance improvement clearly demonstrates the effectiveness of the proposed MMF module.

\begin{figure}[t]
\centering  
\includegraphics[width=0.455\textwidth]{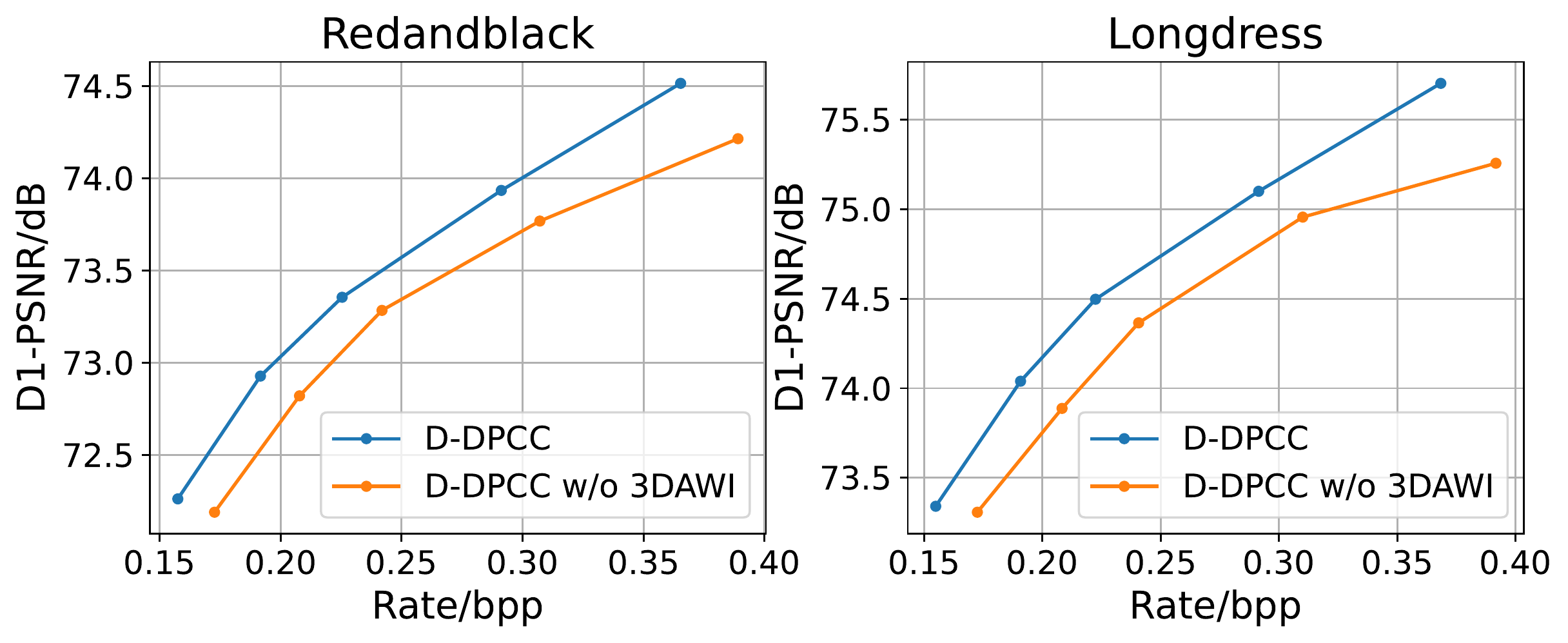}

\caption{Ablation study on Redandblack and Longdress for 3D Adaptively Weighted Interpolation (3DAWI).}
\label{alpha_ablation}
\end{figure}

\paragraph{Effectiveness of 3D Adaptively Weighted Interpolation.}  The objective performance with/without the penalty coefficient $\alpha$ of 3D Adaptively Weighted Interpolation (3DAWI) is presented in Figure \ref{alpha_ablation}. It can be observed that 3DAWI significantly improves the objective performance, with $12.63\%$ D1 BD-Rate gain on Redandblack and $14.33\%$ D1 BD-Rate gain on Longdress. 3DAWI produces a point-wise feature prediction and penalizes those {\it isolated} points that deviate from their 3-nearest neighbors in the referenced point cloud, by adaptively decreasing their weighted sum. Therefore, 3DAWI brings significant improvement in coding efficiency. The minimum distance between points in 8iVFB is 1; thus, $\alpha$ is set as 3 in experiments.

\section{Conclusion}
This paper proposes a Deep Dynamic Point Cloud Compression (D-DPCC) framework for the compression of dynamic point cloud geometry sequence. We introduce an inter prediction module to reduce the temporal redundancies. We also propose a Multi-scale Motion Fusion (MMF) module to extract the multi-scale motion flow. For motion compensation, a 3D Adaptively Weighted Interpolation (3DAWI) algorithm is introduced. The proposed D-DPCC achieves $76.66\%$ BD-Rate gain against state-of-the-art V-PCC Test Model v13.

\section*{Acknowledgement}
This paper is supported in part by National Key R\&D Program of China (2018YFE0206700), National Natural Science Foundation of China (61971282, U20A20185). The corresponding author is Yiling Xu (e-mail: yl.xu@sjtu.edu.cn).
\bibliographystyle{named}
\bibliography{ijcai22}

\end{document}